\pdfoutput=1

\documentclass[11pt]{article}

\usepackage[preprint]{acl}

\usepackage{times}
\usepackage{latexsym}

\usepackage[T1]{fontenc}

\usepackage[utf8]{inputenc}

\usepackage{microtype}

\usepackage{inconsolata}

\usepackage{graphicx}

\usepackage{acl}
\usepackage{float}
\usepackage{bbm}
\usepackage[inkscapeformat=png]{svg}
\usepackage{amsmath}
\usepackage{amssymb}

\usepackage{wrapfig}
\usepackage{tikz}
\usepackage[linguistics]{forest}
\usepackage{amsthm}
\usetikzlibrary{calc}
\usetikzlibrary {positioning}

\usepackage{mathtools}
\usepackage[createShortEnv]{proof-at-the-end}

\newtheorem{theorem}{Theorem}[section]

\newtheorem{definition}[theorem]{Definition}
\newtheorem{example}[theorem]{Example}

\newcommand{\ins}{\vartriangleleft}

\newcommand{\R}{\mathbb{R}}
\newcommand{\N}{\mathbb{N}}

\newcommand{\cF}{\mathcal{F}}
\newcommand{\cI}{\mathcal{I}}
\newcommand{\cT}{\mathcal{T}}

\newcommand{\labelledtree}[1]{
\scalebox{0.8}{
\begin{forest}
  for tree={circle,draw,minimum size=1em,inner sep=0.5pt}
  #1
\end{forest}
}}

\newcommand{\raisetree}[1]{
\raisebox{1em}{#1}
}

\newcommand{\tinytree}[1]{
\raisetree{
\scalebox{0.4}{
\begin{forest}
  for tree={circle,fill,}
  #1
\end{forest}
}
}}

\newcommand{\mediumtree}[1]{
\raisetree{
\scalebox{0.8}{
\begin{forest}
  for tree={circle,fill,}
  #1
\end{forest}
}
}}

\newcommand{\doublenode}{draw,double,double distance=4pt,line width=2pt}
\newcommand{\doublerednode}{draw,red,double,double distance=4pt, line width=2pt}

\forestset{
  downroof/.style={
    for children={
      if n=1{
        edge path'={
          (.parent first) -- (!u.parent anchor) -- (!ul.parent last) -- cycle
        }
      }{no edge}
    }
  }
}
\forestset{
  nice empty nodes/.style={
    for tree={calign=fixed edge angles, calign angle=35},
    delay={where content={}{shape=coordinate,
    for current and siblings={anchor=north}}{}}
  }
}
\usepackage{multicol}

\title{A Lie-algebraic perspective on Tree-Adjoining Grammars}

\author{Isabella Senturia \\ Yale University \\ \texttt{isabella.senturia@yale.edu} 
\And Elizabeth Xiao \\ California Institute of Technology \\ \texttt{exiao@caltech.edu }
\AND Matilde Marcolli \\ California Institute of Technology \\ \texttt{matilde@caltech.edu}}

\begin{document}
\maketitle

\begin{abstract}

We provide a novel mathematical implementation of tree-adjoining grammars using two combinatorial definitions of graphs. With this lens, we demonstrate that the adjoining operation defines a pre-Lie operation and subsequently forms a Lie algebra. We demonstrate the utility of this perspective by showing how one of our mathematical formulations of TAG captures properties of the TAG system without needing to posit them as additional components of the system, such as null-adjoining constraints and feature TAG.

\end{abstract}

\section{Introduction}\label{intro}

Formal languages have long been one of the most prominent tools used by mathematical linguistics in attempts to develop formal systems generating natural language \cite{harrison1978introduction,sipser1996introduction,hopcroft2001introduction}. Different grammars enable descriptions of different phenomena in natural language to varying degrees. While the goal is often conceptualized as generating a given language as string of words, the inherent hierarchical nature of language leads to another aim: to generate a language via the appropriate tree structures modeling those hierarchical relationships within natural language. Tree-Adjoining Grammars  (TAGs) \cite{joshi1975tree,joshi1987introduction,joshi1997tree,kroch1985linguistic} are one approach to such a goal. Structure building via Merge in Minimalism is another approach.

At the same time, mathematicians have worked to reframe various linguistic structures and formulations as mathematical systems. With respect to formal languages, and in particular tree languages, \citet{giraudo2019colored} demonstrates that the algebraic structures known as colored operads can successfully define various string languages (such as context-free languages) and tree languages (including those generated by tree grammars, regular tree grammars, and synchronous grammars). Most recently, \citet{marcolli2025mathematical} have algebraically formalized Merge in the framework of the Strong Minimalist Thesis, modeling syntactic derivations in Hopf-algebraic terms. Also relevant to the Lie algebra formalism considered here,  \citet{marcolli2015graph} use different combinatorial definitions of graphs to show specific context-free and context-sensitive graph grammars have associated Lie algebras.

With respect to linguistic structure, operations to build larger trees from smaller trees are fundamental in capturing the compositional nature of syntax. In Minimalism, two trees are combined together by Merge by creating a new node and two edges, each of which joins to the two respective roots of the other trees. A complimentary perspective to this external composition, where the combining operation does not affect the internal structure of either of the two independent components, would be internal composition, where one tree is inserted \textit{into} another. This operation is exactly the way that TAG operates, and pre-Lie operations' ability to capture this underpins the relevance of Lie algebras to syntactic structure-building. Moreover, there are two complimentary types of insertion of one tree into another, i.e. at a node (as in TAGs) or at an edge.\footnote{ Edge-insertion provides another example of a pre-Lie operation which can be compared and contrasted to the TAG pre-Lie operation. Due to issues of space, we are unable to expand upon this in this work.} This paper explores the pre-Lie operation derived from node-insertion.

One mathematical motivation for studying Lie algebras is their close and richly-studied relationship with Hopf algebras. Every Lie algebra has a Hopf algebra associated to it through its universal enveloping algebra; conversely, the Milnor-Moore theorem provides conditions for a Hopf algebra to occur as the universal enveloping algebra of some Lie algebra. Pre-Lie algebras are also closely tied to rooted trees in that the free pre-Lie algebra generated by a single element is isomorphic to the vector space of nonplanar rooted trees equipped with an insertion pre-Lie operator, which also has an inherent operadic description \cite{chapoton}. In fact, the Connes-Kreimer Hopf algebra of rooted trees is dual to the universal enveloping algebra of the Lie algebra from a free pre-Lie operator \cite{connes1999hopf}.

In this work, we explore the formal properties of tree-adjoining through the mathematical lens of Lie algebras. Although we initially define adjunction on trees in a more general setting, we can constrain our analysis to a particular TAG by restricting the set of generators to the initial and auxiliary trees which occur in that TAG. By recasting TAGs as an algebraic system, we demonstrate that the most appropriate mathematical representation of TAG that yields a Lie algebra requires TAG to be defined using the combinatorial definition of graphs in terms of corollas and half-edges (typically used in theoretical physics but not in computer science or formal language theory, so we refer to it as the ``physics definition"). By allowing a single edge to be described as comprising of two half-edges, this provides a more elegant description of the elementary and auxiliary trees and their compositions. With this physics definition of trees, components of the TAG system such as null-adjoining constraints and specification of insertion positions naturally follow and do not need to be postulated ad-hoc (as they do with standard formulations of TAG). Not only this, but elaborations of TAG such as feature-TAG are easily implementable. 

From the algebraic side, we show that various alternatives to the physics definition (undirected binary tree graphs and an elaboration on these, which we coin \textit{double vertex} trees) cause problems with the Lie algebra structure (in particular, the pre-Lie operation) and also converge to the physics definition as the most natural algebraic description. We also show that colored operads can model TAG insertion via the composition of two insertion operations at the leaf.

A formulation of TAGs in terms of Lie algebra and operad insertions was suggested in Section~2.5.1 of \citet{marcolli2025mathematical}, for the purpose of comparison with Merge, and is implemented here. While in formal languages generative models are usually compared in terms of their weak and strong generative capacity, a more refined comparison is possible, in the sense of being able to more explicitly define and compare the underlying algebraic structures. 

\section{Background}

In this section, we introduce the linguistic and mathematical concepts and tools necessary for the paper. We begin with TAGs, before defining graphs and Lie algebras.

\subsection{Tree-adjoining grammars}

We follow the presentation of tree-adjoining grammars as given in \citet{joshi1997tree}. We define the grammar below, before defining the two types of operations that allow building of larger trees out of insertion/composition of smaller trees.
\begin{definition}
    A tree-adjoining grammar (TAG) is a 5-tuple $(\Sigma,N,I,A,S),$ where 
        $\Sigma$ and $N$ are finite sets of terminal and nonterminal symbols, respectively ($\Sigma \cap N = \emptyset$);
        $S\in N$ is a distinguished nonterminal symbol known as the \emph{start symbol};
        $I$ and $A$ are finite sets of (finite) trees called \emph{initial} and \emph{auxiliary} trees, respectively, whose interior nodes are labeled by nonterminal symbols, leaves of initial trees are labeled by either terminal or nonterminal symbols and those labeled by nonterminal symbols are marked for substitution. The auxiliary trees' leaves are all marked for substitution except one, denoted by an asterisk and required to be labeled with the same label as the root.
\end{definition}
\emph{Lexicalized TAG} requires at least one terminal symbol to appear at a leaf of every initial or auxiliary tree. \emph{Elementary trees} are those contained in $I \cup A$, and \emph{derived trees} are those built by composition of two or more trees. The two composition operations are defined as follows:

\begin{definition}
    The \emph{adjoining} operation builds a new tree from an auxiliary tree $S$ and an initial, auxiliary or derived tree $T$. This is done by inserting $S$ into $T$ at a node $\alpha$ when the root of $S$ and a leaf of $S$ are both labeled with $\alpha$. In other words, the node labeled $\alpha$ in $T$ is replaced by the tree $S$. We denote this operation, somewhat nontraditionally, as $$T \ins_\alpha S,$$ in order to introduce the insertion operation which is relevant for the Lie algebra. 
\end{definition}
\begin{example}\label{init}
Let $S$ and $T$ be the following trees:\footnote{Adapted from \citet{joshi1997tree}.}
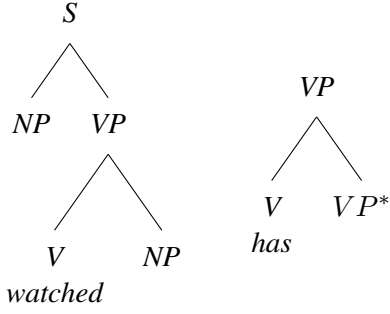
\begin{center}
$\textbf{T:}$ \hspace{2.8cm} $\textbf{S:}$
\end{center}
\begin{center}
\hspace{-3cm}
\begin{forest}{nice empty nodes}
        [S [NP][VP [V \\watched][NP]]]
    \end{forest}    

\vspace{-3.5cm}
\hspace{3cm}
\begin{forest}{nice empty nodes}
        [VP [V \\has][$VP^\ast$]]
    \end{forest}   
\end{center}
\vspace{2cm}
Then inserting $S$ into $T$ at the VP node of $T$ yields
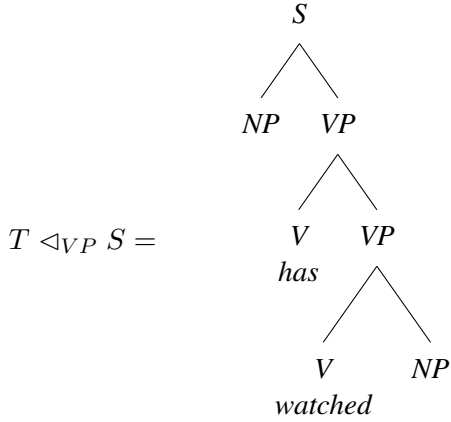
\begin{center}
    
$T \ins_{VP} S=$\hspace{1cm}\raisebox{3cm}{\begin{forest}{nice empty nodes}
        [S [NP][VP [V \\has] [VP [V \\watched][NP]]]]
    \end{forest}  }
    \end{center}

\end{example}

The second operation available in TAGs, \emph{substitution}, was not included in the original definition of TAGs and in fact does not provide any additional expressive power. As such, our Lie-algebraic formalization of TAGs explicitly implementing only the adjoining operation (as the pre-Lie operation) is sufficient to express the entirety of the TAG formalism.

Typically, the arity of TAG trees is unary-binary---that is, nodes can have 0, 1 or 2 children. One can restrict to (full) binary trees (all non-leaves have two children and leaves have zero children) to simplify the set of objects or for comparison to other analyses (Lie-algebras of binary trees and linguistic analyses which consider Merge to be binary, such as \citealp{marcolli2025mathematical}).

TAG trees have labeled insertion spots at interior nodes. In our formulation, one can either have uniform insertions at every interior node, or labeled insertion: both cases are realized as a sum over all possible insertions. Similarly, we allow re-attaching of the remainder (part beneath the insertion spot) of the tree being inserted into to be reattached at any leaf, and denote this as a sum over all leaves. 
We discuss variations of this labeling later in the paper when we demonstrate the ability of the physics graph formulation of TAG to seamlessly incorporate null-adjoining constraints and feature-TAG.

\subsection{Math}

We now present the mathematical concepts utilized in this paper: graphs, Lie algebras, and operads.

\subsubsection{Graphs}\label{graph}

The usual combinatorial definition of graphs is in terms of vertices and edges, namely
we define a graph $G = (V(G),E(G))$, where  $V(G) = \{v_1, v_2,…,v_n\}$ is a set of $n$ \textit{vertices}, $E(G)=\{e_1= (v_a, v_b), \dots, e_m=(v_p, v_q) \}$ is a set of $m$ \textit{edges}. If the edges are undirected, the edge pair $(v_i, v_j)$ is unordered, whereas if the edge is directed, the edge pair is ordered $(start, end)$.

The \textit{degree} $d_v$ of a vertex $v$ is the number of edges connected to that node. A \textit{leaf} is a node of degree 1.
Two \textit{adjacent} vertices are connected by an edge. A \textit{path} from some vertex $v_i$ to another $v_j$ is the sequence of edges connecting adjacent nodes between $v_i$ and $v_j$. A graph is \textit{connected} if there is a path from every node to every other node.

The class of \textit{trees} is the class of connected \textit{acyclic graphs} $T=(V,E)$ defined by the existence of exactly one path connecting any two distinct vertices $v_1, v_2 \in V$---that is, they have no loops. A \textit{directed tree} is a tree with directed edges. A \textit{rooted} tree is a tree for which a specific node has been designated as the root, and is graphed with this root at the top or bottom. We refer to the set of nodes adjacent to, and below, another node $v$ as the children of $v$. Any rooted tree can be viewed as a directed tree, where all edges are either uniformly directed towards the root, or away from it. A tree is \emph{planar} (or, planarly embedded) if for all $v$, the children of $v$ have a linear order.

\subsubsection{Lie algebras}

We follow the definition of Lie and pre-Lie algebras given in \citet{procesi2007lie} and \citet{cartier-patras}. All vector spaces are assumed to be over the real numbers $\R$, although our exposition will work for any field of characteristic $0$.

A \emph{Lie algebra} is a vector space $V$ equipped with a bilinear product $[a,b]$ satisfying the Lie axioms of \emph{antisymmetry}, 
\[ [a,b] = -[b,a], \]
and the \textit{Jacobi identity},
\[ [a,[b,c]] + [b,[c,a]] + [c,[a,b]] = 0.\]
This product is called the \textit{Lie bracket}. Due to antisymmetry, the Lie bracket is not commutative unless it is uniformly zero.

A vector space $V$ is \emph{graded} if it decomposes into a direct sum indexed by the non-negative integers: $V = \bigoplus_{n=0}^\infty V_n$, where each $V_n$ is a subspace of $V$ and  $V_i \cap V_j = \{0\}$ whenever $i\neq j$. We say a graded vector space $V$ is \emph{connected} if $V_0$ is one-dimensional.\footnote{This terminology originates from topology, where the number of generators of the zeroth homology group of a topological space counts its connected components.} If $a \in V_n$, then $a$ is called a \emph{homogeneous} (or \emph{pure}) \emph{element of degree $n$}. If each $V_n$ is finite-dimensional, then $V$ is \emph{locally finite}. 

A Lie algebra $L$ is graded if the Lie bracket respects the grading: if $a \in L_m$ and $b \in L_n$, then $[a,b] \in L_{n+m}$.

\begin{example}
    Any associative algebra $A$ has a Lie bracket defined by $[a,b] := ab-ba$, which is called the \emph{commutator} of $a$ and $b$. For instance, let $A = M_2(\R)$, the space of $2 \times 2$ matrices. Then
    \[ 
        [ \begin{psmallmatrix}1 & 0 \\ 0 & 0\end{psmallmatrix} ,
          \begin{psmallmatrix}2 & 1 \\ 1 & 0\end{psmallmatrix} ] 
        =
        \begin{psmallmatrix}2 & 1 \\ 0 & 0\end{psmallmatrix} 
        -
        \begin{psmallmatrix}2 & 0 \\ 1 & 0\end{psmallmatrix}
        =
        \begin{psmallmatrix}0 & 1 \\ -1 & 0\end{psmallmatrix}.
    \]
\end{example}

We can hence say the following about a commutative algebra:

\begin{example}
    If $A$ is a commutative algebra, then $[a,b] = 0$ for all $a,b \in A$ and the commutator Lie bracket is trivial.
\end{example}

More generally, the Lie bracket is the commutator of a binary operation (such as a product or a pre-Lie operation), meaning it can be considered to be a measure of the degree to which the operation is not symmetric. 

Let $L, M$ be two Lie algebras. A linear map $\phi : L \to M$ is called a \emph{Lie algebra homomorphism} if it commutes with the Lie brackets; that is, for $a,b \in L$,
\[ \phi([a,b]_{L}) = [\phi(a), \phi(b)]_{M}. \]

\subsubsection{Pre-Lie algebras}

Let $V$ be a vector space. A bilinear product $\ins : V \times V \to V$ is called a \emph{(right) pre-Lie operator} if it satisfies the \emph{(right) Vinberg identity}
\begin{align*}
   (a \ins b) \ins c &- a \ins (b \ins c) \\
  &=  (a \ins c) \ins b - a \ins (c \ins b),
\end{align*}
making $(V,\ins)$ a \emph{(right) pre-Lie algebra}. The expression $(a \ins b) \ins c - a \ins (b \ins c)$ is referred to as the \emph{associator} $A(a,b,c)$ with respect to $\ins$; if $\ins$ is associative, then $A(a,b,c) = 0$ uniformly. The Vinberg identity asserts that the associator is unchanged when the second and third arguments are swapped; that is, for all $a,b,c \in V$,
\[ A(a,b,c) = A(a,c,b). \]

A pre-Lie operator induces a Lie bracket defined by $[a,b] := a \ins b - b \ins a$, which automatically satisfies antisymmetry and the Jacobi identity, making $A$ a Lie algebra. We do note that not all Lie algebras arise from a pre-Lie operator.

\subsubsection{Tree-insertion as a free pre-Lie operator}

Let $V={\rm span}(B)$ be a vector space, say of countable dimension, with basis given by $B = \{a,b,c,\dots\}$. We use ${\rm span}(B)$ and $\langle B \rangle$ interchangeably to refer to the span. The \emph{free associative algebra over $V$} has, as its basis, words over the alphabet $B$, such as $aabc$, $acccbba$ or the empty word $\varnothing$; the product is defined over basis elements by concatenation of words with no other relations. This construction is often denoted by $T(V)$ and called the \emph{tensor algebra} over $V$; if $V$ is $d$-dimensional, we often call $T(V)$ the \emph{free associative algebra on $d$ generators} (or letters), which is unique up to isomorphism. Any tensor algebra has a natural grading over $\N$, where the $n$-degree component of $T(V)$ is generated by words of length $n$. There is an injective copy of $V$ embedded in $T(V)$, and any extension of $V$ to an associative algebra is a quotient of $T(V)$.

We can similarly define $L(V)$, the \emph{free pre-Lie algebra over $V$}, with pre-Lie operator $\ins$ satisfying the Vinberg identity on the associators and no other relations. If $V={\rm span}(B)$ is $d$-dimensional, then we will refer to $L(V)$ as a \emph{free pre-Lie algebra on $d$ generators} (or letters), the elements of $B$. However, words over $B$ are no longer sufficient to form a basis for $L(V)$, since $\ins$ is not associative; instead, we require strings to be parenthesized such that each pair of parentheses corresponds to one application of the $\ins$ operator. One interpretation would be as planar full binary trees with leaves labeled by elements of $B$, since such trees with $n$ leaves are in correspondence with complete parenthesizations of a string of length $n$. This basis works for any non-associative algebra over $V$ with no further relations, and it is reminiscent of the Merge operation since it combines left and right arguments into a single tree by grafting them together as the left and right child, respectively, of a new root node. However, this basis does not provide us with a convenient geometric or combinatorial interpretation of the associator, since they are defined over a difference of basis elements.

Another interpretation of the free pre-Lie algebra over $V$ is due to \citet{chapoton}, where basis elements are nonplanar rooted trees (not necessarily binary), with vertices labelled by basis elements of $V$.
The pre-Lie operator can now be interpreted geometrically as insertion: if $T$ and $S$ are two trees, then $T \ins S$ is the sum of all possible trees that result from grafting $S$ to $T$ by joining a node of $T$ to the root of $S$ with a new edge.

Since a new edge is produced in every nontrivial insertion, trees are graded by numbers of vertices and $\ins$ respects the natural grading. The degree zero component of $L(V)$ is generated by the empty tree $\varnothing$; for any tree $T$, we have $T \ins \varnothing = T$ and $\varnothing \ins T = \varnothing$, which extends to the entire vector space by linearity. In the following example, note that the trees are non-planar and so we are currently ignoring linear ordering of the leaves.

\begin{example}
  Let 
  \[ T = \raisetree{\labelledtree{[a[b][c]]}}, \quad S = \labelledtree{[b]}. \]
  Then
  \[
    T \ins S =
    \raisetree{\labelledtree{[a[b[b]][c]]}}
    +
    \raisetree{\labelledtree{[a[b][c[b]]]}}
    +
    \raisetree{\labelledtree{[a[b][c][b]]}}.
  \]
\end{example}

This model provides a geometric interpretation for the associator. If $T_1, T_2, T_3$ are three trees, then $(T_1 \ins T_2) \ins T_3$ is a sum of all trees obtained by first inserting $T_2$ into $T_1$, and then inserting $T_3$ into the resulting tree. The associator consists exactly of the trees produced by inserting $T_2$ and $T_3$ into distinct vertices of $T_1$.

Fix a label $\alpha \in B$ from the basis. We can define a restricted insertion operator $\ins_\alpha$ where grafting only occurs at nodes in the first argument labelled $\alpha$; this is still a pre-Lie operator. The general pre-Lie operator allowing insertion at every node is the sum of the restricted operator over all basis elements: $\ins \, = \sum_{\alpha \in B} \ins_\alpha$.

The insertion operators here will serve as a framework for our definition of TAG as a pre-Lie operator. Tree-adjoining will be viewed as a kind of ``insertion'' where, instead of grafting one tree to a node of the other, the inserted tree becomes embedded inside another one.

\subsubsection{Tree-adjoining as a pre-Lie operator}

Let $X$ be a collection of trees. 
For now, we will assume the trees in $X$ are binary, planar and unlabelled. Let $\cT = \cT_{bin,pl}$ be the free vector space with $X$ as its basis. We define $\ins$ on basis elements $T,S \in X$ by adjoining $S$ at every possible position in $T$ over every leaf of $S$; then, $T \ins S$ is the sum of all trees produced by this process.

Note that we permit adjunction of $S$ into $T$ at leaves of $T$. This procedure superficially resembles the process of substitution, where a leaf is replaced by a larger tree. However, adjunction at leaves is different because it implicitly pairs the root of the inserted tree with one of its leaves, whereas substitution only considers the root of the inserted tree. Hence the coefficient of a tree obtained in such a way in $T \ins S$ will be at least the number of leaves in $S$.

In general, a tree having a coefficient larger than $1$ in $T \ins S$ means it is obtainable from adjunction in multiple different ways.

If $T$ has $n$ vertices and $S$ has $\ell$ leaves, then the sum of the coefficients of $T \ins S$ will be $n\ell$. The operation $\ins$ is then extended to the whole vector space $\cT$ by linearity in both arguments.

\begin{example}
Let
\[ T = \tinytree{[[][[][]]]} \quad \text{ and } \quad S = \tinytree{ [[][]] }, \]
which respectively have 5 nodes and 2 leaves.  We show the result of adjoining $S$ to $T$. The colors are only used for illustrative purposes to highlight the  copy of $S$ in each term; the actual trees are unlabelled.
\begin{align*}
T \ins S
&=
\tinytree{[,fill=red[,fill=red,edge=red[][[][]]][,fill=red,edge=red]]}
+
\tinytree{[,fill=red[,fill=red,edge=red][,fill=red,edge=red[][[][]]]]}
+
\tinytree{[[][,fill=red[,fill=red,edge=red[][]][,fill=red,edge=red]]]}
+ \tinytree{[[][,fill=red[,fill=red,edge=red][,fill=red,edge=red[][]]]]}\\
&\quad\quad
+2 \tinytree{[[,fill=red[,fill=red,edge=red][,fill=red,edge=red]][[][]]]}
+2 \tinytree{[[][[,fill=red[,fill=red,edge=red][,fill=red,edge=red]][]]]}
+2 \tinytree{[[][[][,fill=red[,fill=red,edge=red][,fill=red,edge=red]]]]} \\
&= 
\tinytree{[[[][[][]]][]]}
+
4\tinytree{[[][[][[][]]]]}
+
3\tinytree{[[][[[][]][]]]}
+2 \tinytree{[[[][]][[][]]]}
\end{align*}
In the first expression for $T \ins S$, the last three terms each have a coefficient of $2$ because adjunction occurs at a leaf of $T$. There is no subtree to be attached to any leaf of $S$; however, a leaf of $S$ still needs to be chosen to ensure consistency in the associator, so the two copies correspond to the choices of leaf in $S$.

We reiterate that each tree has a fixed planar embedding; indeed, if the trees were nonplanar, then the first three terms in the last line would represent the same tree.

\end{example}

The insertion operation above has the following property:

\begin{theoremE}[][end, restate]
  $\ins$ is a pre-Lie operator.
\end{theoremE}

\begin{proofE}
  It is enough to show that the right Vinberg identity holds for basis elements. Let $T_1, T_2, T_3$ be trees in $X$. In the associator $A(T_1,T_2,T_3) =  (T_1 \ins T_2) \ins T_3 - T_1 \ins (T_2 \ins T_3)$, the only basis elements with nonzero coefficient are the ones where $T_2$ and $T_3$ are adjoined at distinct vertices of $T_1$, meaning that $T_2$ and $T_3$ are disjoint subgraphs of the resulting tree. These terms will appear in $(T_1 \ins T_2) \ins T_3$, but not $T_1 \ins (T_2 \ins T_3)$. Any term where $T_3$ is adjoined to $T_2$, or a copy of $T_2$ in a term of $T_1 \ins T_2$, necessarily appears in both terms of the associator, hence are cancelled out. This cancellation ensures that $A(T_1, T_2, T_3) = A(T_1, T_3, T_2)$.
\end{proofE}

\begin{example}
  Let 
  \[ T_1 = \tinytree{ [[][]] }, \quad T_2 =  {\color{blue}\tinytree{ [[][[][]]] }}, \quad T_3 =  {\color{red}\tinytree{ [[][]] }}. \]
  Then \[ \tinytree{[,fill=blue[,fill=blue,edge=blue][,fill=blue,edge=blue[,fill=blue,edge=blue[][,fill=red[,fill=red,edge=red][,fill=red,edge=red]]][,fill=blue,edge=blue]]]} \]
  appears as a term of $(T_1 \ins T_2) \ins T_3$ only, since adjunctions occur at different vertices of $T_1$, while
  \[ \tinytree{[  [,fill=blue[,fill=blue,edge=blue][,fill=red,edge=blue[,fill=red,edge=red][,fill=red,edge=red[,fill=blue,edge=blue][,fill=blue,edge]]]] []] } \]
  appears in both $(T_1 \ins T_2) \ins T_3$ and $T_1 \ins (T_2 \ins T_3)$, hence will be cancelled out in the associator.
\end{example}

An immediate question is whether or not tree-adjoining, in any formulation (planar or non-planar, labelled or unlabelled), is isomorphic as a pre-Lie algebra to a free one. This question is relevant to an algebraic comparison between TAGs and the insertion Lie algebra dual to the Hopf algebra of workspaces in Minimalism. Before we can answer this question, we will need to fix some more conditions on the basis elements of $\cT$ and how exactly the tree-adjunction operation is defined. Several different possibilities will be considered in \S \ref{mathform} on the mathematical formulation of TAG, but we will use the binary planar unlabelled case as our first example.

We note the behaviour of the single-node tree in insertions:
\begin{example}\label{ident}
  Let $T$ be a binary planar unlabelled tree, and let $\bullet$ be the tree consisting of a single node. Then
  \begin{align*}
    T \ins \bullet &= |T| T \\
    \bullet \ins T &= \ell(T) T
  \end{align*}
  where $|T|$ is the number of nodes in $T$, and $\ell(T)$ is the number of leaves. Hence the Lie bracket of $T$ and $\bullet$ is
  \[ [T,\bullet] = (|T| - \ell(T)) T. \]
\end{example}

\begin{theoremE}[][end, restate]
  $\cT_{bin,pl}$ is not isomorphic as a pre-Lie algebra to $L_{free}$, the free pre-Lie algebra on one generator.
\end{theoremE}

\begin{proofE}
  Suppose, for the sake of contradiction, that $\cT_{bin,pl}$ is isomorphic to $L_{free}$; then there exists an isomorphism $\phi : L_{free} \to \cT_{bin,pl}$ such that $\phi(x \ins y) = \phi(x) \ins \phi(y)$ for all $x,y \in L_{free}$. 
  (We use the same symbol $\ins$ to denote the pre-Lie operation on both algebras, since context prevents any ambiguity.)
  
  The contradiction is defined by considering the inverse element $\phi^{-1}(\bullet) \in L_{free}$, and showing that it cannot be well-defined.

  We will use the basis on $L_{free}$ given by unlabelled nonplanar trees of arbitrary arity, where trees are graded by its number of nodes.
  
  Since $L_{free}$ is graded, $\cT_{bin,pl}$ inherits a grading from the isomorphism, although it may not be an obvious one based on any combinatorial properties of trees. In fact, a basis element of $\cT_{bin,pl}$ may not even be a homogeneous element under this grading.

  Let $T$ be a tree in $\cT_{bin,pl}$; then $T \ins \bullet = |T|T$, so
  \[ \phi^{-1}(T) \ins \phi^{-1}(\bullet) = |T| \phi^{-1}(T). \]
  
 Let $c \in \R$ be the coefficient of $\varnothing$ in $\phi^{-1}(\bullet)$, so that
  \[ \phi^{-1}(\bullet) = c \varnothing + (\text{higher degree terms}). \]
  
   Let $y$ be a term in $\phi^{-1}(T)$ of minimal degree. Since $\varnothing$ is the unique degree-0 generator in $L_{free}$, it is the only element that does not raise the degree of $y$ when inserted into it. Since $|T| y$ occurs in the expression $\phi^{-1}(T) \ins \phi^{-1}(\bullet)$, it can only be the result of performing $\phi^{-1}(T) \ins c\varnothing$, hence $c = |T|$. But $\phi^{-1}(\bullet)$ cannot depend on any specific $T$.
\end{proofE}

To prevent overgeneration, we can introduce labels to $\cT_{bin, pl}$ and modify the adjunction operation so that adjunction only occurs when an insertion spot of $T$ has the same label as the root of $S$, and moreover that $S$ has at least one leaf with the same label, to which the subtree of $T$ will be attached. If $S$ has a unique such leaf, then it represents the distinguished foot node of an auxiliary tree in TAG. See \ref{1graph} for an example.

\subsubsection{Operads}

As mentioned in the introduction, operads have recently been connected to mathematical linguistics \cite{giraudo2019colored,marcolli2025mathematical}. Colored operads have most recently been used in formal language theory to obtain a new proof of the well-known Chomsky-Sch\"utzenberger representation theorem \cite{mellies2023categorical}. Colored operads are also used in \citet{ml2025theta} and \citet{mhl2025phases} to model theta roles and phases in Minimalism, as outlined in \citet{marcolli2025mathematical}. Indeed, in this work we also utilize colored operads as operations on the vector space containing TAG trees that can mimic the TAG pre-Lie insertion operation as a combination of two colored-operad compositions. We define the notion of a colored operad below, before returning to it in section \ref{operadictag} to demonstrate how it appropriately captures the Lie-algebraic TAG system.

We provide here a preliminary introduction to the theory of colored operads, and we refer the reader to \citet{giraudo2019colored} for a more detailed formulation in a formal languages context.

An \emph{operad} organizes the compositional structure of operations that take multiple inputs and produce a single output. It is a collection $\mathfrak{D}$ of sets $\mathfrak{D}(n)$, $\mathfrak{D} = \lbrace \mathfrak{D}(n)\rbrace$ that consist of operations $T \in \mathfrak{D}(n)$ with $n$ inputs and one output. The operad has an algebraic structure defined by composition operations 
\[ \gamma : \mathfrak{D}(n) \times \mathfrak{D}(k_1) \times ... \times \mathfrak{D}(k_n) \rightarrow \mathfrak{D}(k_1+...+k_n) \]

These composition operations take the single output out of an operation in $\mathfrak{D}(k_i)$ and plug it into the $i$-th input of an operation in $\mathfrak{D}(n)$. This results in a new operation in $\mathfrak{D}(k_1+...+k_n)$ whose set of inputs is the union of all the inputs of the operations $\mathfrak{D}(k_i)$, and a single output, which is the output of the operation in $\mathfrak{D}(n)$ they are composed with. The compositions $\gamma$ are associative. Operads can be \emph{unital}, \emph{symmetric}, etc., but we omit these definitions here as they are not relevant for the current work.

An \emph{algebra $A$ over an operad $\mathfrak{D}$} is a set that can serve as inputs and outputs for the operations in $\mathfrak{D}$, namely for which there are maps
\[ \gamma_A : \mathfrak{D}(n) \times A^n \to A \] compatible with the $\gamma$ compositions of $\mathfrak{D}$.

A \emph{colored operad} $\mathcal{O} = \lbrace \mathcal{O}(c,c_1,...,c_n)\rbrace$ is an operad where inputs and outputs are labelled (by a set of colors $c, c_i \in \Omega$) and composition can only occur when the output color $c$ of one operation matches the color $c_i$ of the $i$-th input it feeds to.  

In fact, the concept of a pre-Lie operator is inherently operadic. \citet{chapoton} explain that pre-Lie algebras naturally arise as algebras over a binary quadratic operad.

In the next section, we mathematically define TAGs in three different ways, and show that only the third mathematical formulation of TAG (using the physics definition of graphs) suffices when we would like to require the TAG to be a Lie algebra.

\section{Mathematical formulation of TAG}\label{mathform}

To begin formulating TAG as a mathematical system, we use the combinatorial graph definition of trees as given in section \ref{graph}.

\subsection{Normal single vertex method}
 In what follows, we will demonstrate insertion at a specific (labeled) location in the tree, instead of giving the sum over all possible insertions. We label the spot we insert at in $T$, as well as the two locations (root and specified leaf) that the adjoining occurs at in $S$, with $\alpha$. Note also that usually we would be summing over all leaves for the lower part of $T$ to re-attach to, but again for simplicity we omit this.

\begin{example}\label{1graph}
We formulate the trees from the initial TAG example (\ref{init}) as graphs. Let
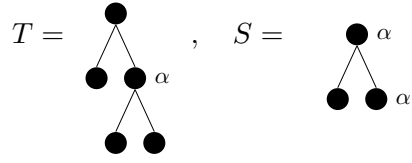
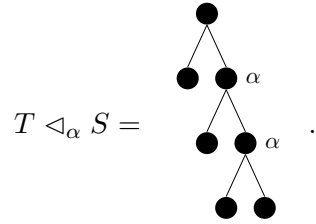
\[
 T =  \mediumtree{ [[] [,label={east:$\alpha$} [][ ]]]}, \quad S =  \raisebox{-\height/2}{ \mediumtree{ [,label={east:$\alpha$}[][,label={east:$\alpha$}]]} } \]

Then
\[ T \ins_\alpha S = \raisebox{1.2cm}{\mediumtree{ [[] [,label={east:$\alpha$} [][,label={east:$\alpha$} [] []]]]}}. \]

\end{example}

The problem with this comes when we think about the grading of the vector space. We are motivated to introduce a grading to our vector space because trees are fundamentally combinatorial objects with a notion of size. Furthermore, provided that the set of nodes labels is finite, then the vector space will be locally finite. Ideally we would also want the degree-0 part of the vector space to be generated by a single element, such as the empty tree, so that the vector space is connected, much as in the case of the free pre-Lie algebra. Indeed, if $V$ is connected as a graded vector space, then the universal enveloping algebra of $L(V)$ is a graded connected Hopf algebra, which carry many desirable properties. The pre-Lie operator and Lie bracket with respect to the degree-$0$ generator should be simple enough that their outputs can be described in closed-form, potentially allowing us to determine whether or not the (pre-)Lie algebra is isomorphic as (pre-)Lie algebras to another one.

If we grade over vertices, we see that the insertion operation does not preserve gradation: inserting an $n$-noded tree $S$ into an $m$-noded tree $T$ yields an $n+m-1$-noded tree, because $S$ is replacing a vertex in $T$ and hence one vertex overall is lost. 

A possible repair to this, to preserve the gradation of the vector space, would be to grade over number of edges rather than number of vertices. Then inserting an $n$-edged tree $S$ into an $m$-edged tree $T$ yields an $n+m$-edged tree, which fixes that problem, but there is now a new issue: there are two unique trees with zero edges. These are the empty tree, and the tree comprising of a single node. The situation is further complicated when trees are labeled, since there is now a distinct single-noded tree for every label. As stated in \ref{ident}, the insertion of any of these single-noded trees will return $T$ multiplied by a factor of $|T|_\alpha$, the number of times $\alpha$ occurs in $T$, meaning it is not exactly the identity insertion operation and hence even more problematic.

Mathematically, we can solve this ambiguity by equating all single-noded trees with the generator of the degree-0 component, which is achieved by quotienting out the ideal generated by elements $1-\bullet_\alpha$ for all labels $\alpha$, where $\bullet_\alpha$ is the single-noded tree where the only node is labelled $\alpha$. However, this trivializes the Lie bracket with respect to $\bullet_\alpha$, and insertions of the form $T \ins \bullet_\alpha$ are no longer eligible to be used as a way of counting occurrences of $\alpha$ in $T$. Thus, we are motivated to reformalize the TAG tree graphs in a different way. This is done with the double vertex method in the following section.

\subsection{Double vertex trees}

\citet{rambow2001d} suggest that locations within a tree where adjunction was possible, which are normally represented as a node in the tree, are actually two nodes, an upper copy and a lower copy, connected by a dashed line (see figure 1).

\begin{wrapfigure}{r}{0.18\textwidth}
  \begin{center}
 \begin{forest}
  [S [NP\\she] [VP[VP, edge = {dashed} [V \\saw][S]]]]
\end{forest}
\end{center}
  \caption{A tree adapted from \citet{rambow2001d} which can be inserted into at the VP node(s).}
  \label{xbar}
\end{wrapfigure}
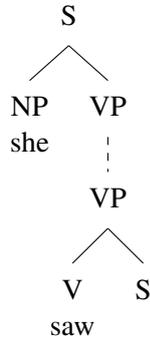

Inspired by this and in order to fix the grading issue raised by defining trees as regular graphs, we can introduce a modification on the previous trees which we call \emph{double-vertex trees}. These trees are also graphical trees, with the caveat that any node $\alpha$ that is able to be inserted into is actually a double node, i.e. is counted as two nodes. The maximum count of any vertex is two, meaning we cannot increase the count of a vertex to any $n \in \mathbb{N}$. In order to prevent this unbounded increase, and because we are working with unlabeled trees, i.e. where adjoining can occur freely at any location, this means that in any non-auxiliary tree every node will be a double vertex. In every auxiliary tree, there will be exactly two single vertices: the root, and one leaf. These are the two nodes that will each get one additional count from the double vertex $\alpha$ they are inserting into, meaning that the root and that distinguished leaf will become double-vertex interior vertices in the larger post-adjunction tree.

We can see this in the following example, which recasts example (\ref{1graph}) in terms of this new counting system. 

\begin{example}
Let $S,T$ be as above, i.e.
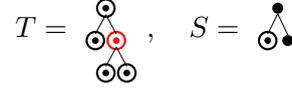
\[ T = \tinytree{[,\doublenode[,\doublenode][,\doublerednode,[,\doublenode][,\doublenode]]]}, \quad  S = \tinytree{[[,\doublenode][]]} \]
Note that every node in $T$ is doubled, and that the color is only used to denote the insertion spot. Inserting $S$ into  $T$ at the red node yields
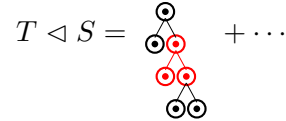
\[ T \ins S = \tinytree{[,\doublenode[,\doublenode][,\doublerednode[,\doublerednode,edge=red][,\doublerednode,edge=red[,\doublenode][,\doublenode]]]]} + \cdots \]

\end{example}

However, this model also has its drawbacks. In particular, we are now able to grade by vertices, because we see that the insertion operation does preserve gradation: inserting an $n$-noded tree $S$ (with $(n-2)/2$ double-nodes and 2 single nodes) into an $m$-noded tree $T$ yields an $n+m$-noded tree (as the double node in $T$ where they are both inserted contributes one node to each of the two single nodes, where adjunction takes place, turning those two single nodes into two double nodes). That being said, there is still a strange effect that this has on the overall gradation of the vector space. All odd-degree components are ignored when restricting to the trees we use in TAG: since all binary trees have an odd number of nodes, odd-degree indicates that there are an odd number of single nodes.

The following is the set of all gradations of the space up to five nodes.

\begin{example}
  \begin{align*}
    \cT_0 &= \langle \varnothing \rangle \\
    \cT_1 &= \langle \tinytree{ [] } \rangle \\
    \cT_2 &= \langle \tinytree{ [,\doublenode] } \rangle \\
    \cT_3 &= \langle \tinytree{ [[][]] } \rangle \\
    \cT_4 &= \langle
      \tinytree{ [,\doublenode[][]] },
      \tinytree{ [[,\doublenode][]] },
      \tinytree{ [[][,\doublenode]] }
    \rangle \\
    \cT_5 &= \langle
      \tinytree{ [,\doublenode[,\doublenode][]] },
      \tinytree{ [,\doublenode[][,\doublenode]] },
      \tinytree{ [[,\doublenode][,\doublenode]] },
      \tinytree{ [[[][]][]] }, 
      \tinytree{ [[][[][]]] }
    \rangle
  \end{align*}
\end{example}

We notice that gradation by nodes results in two trees being able to have the same dimension (number of nodes) but they can have different numbers of edges, e.g. in $\mathcal{T}_5$ there are trees with either two or four edges. This vector space contains many trees that will never be used in tree-adjunction, so we will consider the subspace $\cT' \subseteq \cT$ restricted to the trees that appear in TAG, which is closed under $\ins$. Under the grading inherited from $\cT$, the even-degree components of $\cT'$ have the following generators; the odd-degree, as mentioned above, are trivial.

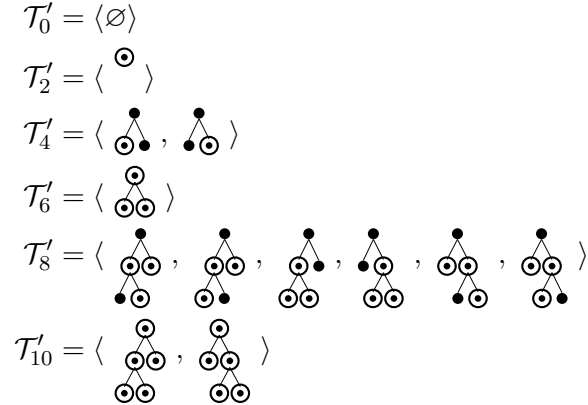
\begin{example}
  \begin{align*}
    \cT'_0 &= \langle \varnothing \rangle \\
    \cT'_2 &= \langle \tinytree{[,\doublenode]} \rangle \\
    \cT'_4 &= 
    \langle
      \tinytree{ [[,\doublenode][]] },
      \tinytree{ [[][,\doublenode]] }
    \rangle \\
    \cT'_6 &=
    \langle
      \tinytree{ [,\doublenode[,\doublenode][,\doublenode]] }
    \rangle \\
    \cT'_8 &=
    \langle
      \tinytree{[[,\doublenode [] [,\doublenode] ] [,\doublenode]]},
      \tinytree{[[,\doublenode [,\doublenode] [] ] [,\doublenode]]},
      \tinytree{[[,\doublenode [,\doublenode] [,\doublenode] ] []]},
      \tinytree{[[] [,\doublenode[,\doublenode] [,\doublenode] ] ]},
      \tinytree{[[,\doublenode] [,\doublenode[] [,\doublenode] ] ]},
      \tinytree{[[,\doublenode] [,\doublenode[,\doublenode] [] ] ]}
    \rangle \\
    \cT'_{10} &=
    \langle
      \tinytree{[,\doublenode[,\doublenode [,\doublenode] [,\doublenode] ] [,\doublenode]]},
      \tinytree{[,\doublenode[,\doublenode] [,\doublenode[,\doublenode] [,\doublenode] ] ]}
    \rangle
  \end{align*}
\end{example}
There is now an alternation in the degrees; the ones which are 0 mod 4 contain the auxiliary trees, whereas the ones which are 2 mod 4 contain the non-auxiliary trees where all nodes are doubled. The difference between the two fundamental types of trees is encoded in the grading. Note that $\cT'$ is still not a free pre-Lie algebra since certain insertions of basis elements produce zero: namely, any attempt to adjoin a non-auxiliary tree into another tree is zero since it is an invalid insertion.

This method's particular utility in being able to encode the different types of trees (auxiliary vs. elementary) in the grading scheme is one reason that these tree graph encodings should continue to be studied in future work. However, the amount of hoops necessary to jump through in order to establish this class of graphs as useful as a mathematical formulation of TAG motivates us to look at a third graphical encoding of TAG trees, which is even simpler than this version and has particularly interesting linguistic implications.

\subsection{TAG as physics graphs}

We present an alternative combinatorial definition of graphs that is primarily used in theoretical physics \citep{marcolli2015graph}, which is  convenient for defining the colored operads model of tree operations.

\subsubsection{Physics definition}

 A graph $G = (C(G), \cF(G), \cI)$ consists of a set $C(G) = \{v_1, \dots, v_n\}$ of $n$ \emph{corollas} (nodes distinguished by having half-edges attached to them), a set $\cF(G) = \{f_1, \dots, f_m\}$ of $m$ \emph{flags} (or \emph{half-edges}), and an involution $\cI : \cF(G) \to \cF(G)$ on the flags. The \emph{external edges} $E_{ext}(G)$ of $G$ are the flags $f \in \cF(G)$ fixed by the involution, i.e. $\cI(f) = f$, while the \emph{internal edges} $E_{int}(G)$ correspond to the two-element subsets $\{f, f'\} \in {\cF(G) \choose 2}$ such that $\cI(f) = f'$ (and correspondingly $\cI(f') = f$). The \emph{valence} of a corolla $v \in C(G)$ is the number of half-edges attached to it.

Operadic insertion is now represented by joining external edges; adjunction involves splitting an internal edge by dividing it into its component flags and joining two external edges of a new graph, one to each flag.

This model also provides a concrete distinction between non-auxiliary trees and auxiliary trees. Auxiliary trees have exactly two half edges; one corresponding to the root, and another corresponding to the leaf that has the same label as the root in Joshi's original formulation of TAG.\footnote{See, for instance, tree $S$ in example (\ref{halfedge}).}

In fact, there is some nuance here. Because we are representing the pre-Lie insertion operation as a summation over all possible insertions, this will be generalized as \textit{all} leaves of auxiliary trees being half edges---the insertion operation of $T \ins S$ then varies across the half edge beneath the insertion point in T being re-attached in turn to each half edge leaf of S. This naturally motivates the colored-operad lens on the TAG Lie algebra---every single leaf is a location where the colored operad will, in turn, insert a terminal node with a half edge attached to it. We return to this idea later on in this section, but note that we will not be writing the trees with terminal symbols and half-edges for all leaves, for sake of convenience.

The concept of half-edges is reminiscent of the usage of half-arcs to represent features in Minimalist grammars, where matching features are joined together. We explore this interpretation further in Section \ref{lingimp}.

\subsubsection{Operadic definition of TAG Lie algebra}\label{operadictag}

The following describes how TAG insertion can be implemented via the composition of two unary operad insertions.

\begin{definition}
    Consider the insertion $T\ins S$, where insertion location is labeled with $\alpha$. $T$ was originally derived from composing two trees at $\alpha$ via an operad composition:
\begin{equation}\label{circalpha}
    T = \mathcal{O}(T_1,T_2) = T_1 \circ_{\alpha} T_2.
\end{equation}
    This $\alpha$ composition location is exactly where $T$ will be broken apart in order to insert $S$ between the two:
    \[ T \ins_\alpha S = \mathcal{O}(\mathcal{O}(T_1,S),T_2) = (T_1  \circ_{\alpha_{u}} S) \circ_{\alpha_{l}} T_2 \]
    $u$ and $l$ stand for ``upper" and ``lower" copies of $\alpha$, respectively.\footnote{One may view conceptualizing a tree containing an insertion spot at an interior node as being comprised of two trees who each contain that node on their periphery (the upper copy being a leaf in $T_1$ and the lower copy being the root of $T_2$) as linguistically-alternative, but any tree can be viewed as the composition of a subgraph and the quotient graph over the subgraph.} Note that $\alpha_u$ actually comes from $S$, and $\alpha_l$ comes from $T$, because the edge above $\alpha$ in $T$ is what is split into two half edges in order to insert $S$.
    
\end{definition}

 Because one of the $\alpha$ nodes is coming from the auxiliary tree $S$, and the other is coming from the tree being adjoined into, $T$, this results in the grading by vertices is preserved with this formalization of graphs, maintaining the connectedness of the space and hence solving those issues as they pertained to the previous two graph formalisms.

In the following example, we demonstrate the physics graph definition as it is applied to the TAG trees, using our running example, and show how the adjoining operation can be performed via the composition of two operad operations.

\begin{example}\label{halfedge}
   An example of the insertion operation (outside of the operad). Supposed the grading is by nodes.

    Let $S,T$ as in the previous examples:
    $$T =\scalebox{.8}
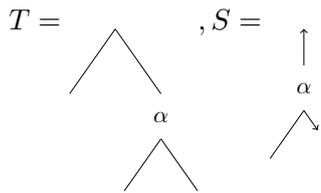
{\begin{forest}for tree=nice empty nodes
        [ [] [$\alpha$[, l*=1.5] [, l*=1.5]]]
    \end{forest}},   S = \scalebox{.8}{\begin{forest}for tree=nice empty nodes
        [[$\alpha$, edge={<-} [][, edge={->, shorten >=1.5em}] ]]
    \end{forest}}$$
To insert $S$ into $T$ at $\alpha$, forming $T \ins_{\alpha} S$, we break $T$ into $T_1$ and $T_2$ such that $T = T_1 \circ_{\alpha} T_2$, and specifically the edge above $\alpha$ in $T$ is broken into two half edges. Hence, $T_2$ retains $\alpha$:

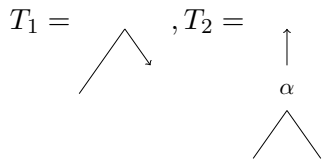
$$ T_{1} = \scalebox{.8}{\begin{forest}for tree=nice empty nodes
        [ [] [,edge={->, shorten >=1.5em} ]]
    \end{forest}}, T_2 = \scalebox{.8}{\begin{forest}for tree=nice empty nodes
        [[$\alpha$, edge={<-}  [] []]]\end{forest}}$$

Then inserting $S$ into $T$ at $\alpha$ is simply the composition 
$$T \ins_{\alpha} S = T_1 \circ_{\alpha} S \circ_{\alpha} T_2$$
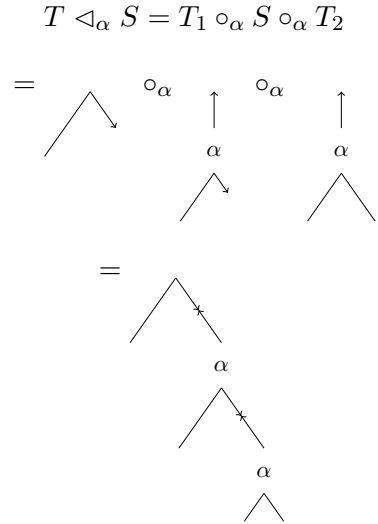
    $$ = \scalebox{.8}{\begin{forest}for tree=nice empty nodes
        [ [] [,edge={->, shorten >=1.5em} ]]
    \end{forest}} \circ_{\alpha}  \scalebox{.8}{\begin{forest}for tree=nice empty nodes
        [[$\alpha$, edge={<-} [][, edge={->, shorten >=1.5em}] ]]
    \end{forest}} \circ_{\alpha} \hspace{.2cm}\scalebox{.8}{\begin{forest}for tree=nice empty nodes
        [[$\alpha$, edge={sloped, <-}  [] []]]\end{forest}}$$

        $$ = \scalebox{.8}{\begin{forest}for tree=nice empty nodes
        [,name = d2 [] [,l*= .5, edge = {->}[,no edge][ $\alpha$, name = d1,l*= .5, edge = {<-} [][,l*= .5, edge = {->} [,no edge][$\alpha$,l*= .5, edge = {<-} [] []] ]  ]]]
    \end{forest} }$$

\end{example}

There is a distinction that must be made regarding the arity of the colored operads we will use in the context of the TAG Lie algebra. In the Lie algebra formalism we sum over insertions at all possible locations, and readjoining at any possible leaf.
Then, in order to complete the TAG tree, after readjoining at one of the leaves, the remaining $l-1$ leaves must be ``capped off" with operadic insertion of $l-1$ half-edges whose single nodes are labeled with terminals. When we use operadic compositions to describe the insertion operations, we can decompose the composition $\gamma$ (that fills all inputs at once) into repeated compositions $\circ_i: \mathfrak{D}(n)\times \mathfrak{D}(m)\to \mathfrak{D}(n+m-1)$ that perform a single match of output to input. When the color-matching is also taken into account, these give \eqref{circalpha}. Thus, we can assume that only one leaf is a half-edge and all the other leaves are filled with terminals. In operadic terms, filling with terminals is part of the algebra over an operad structure, see \citet{giraudo2019colored}.

\section{Implications of mathematical formalizations of TAG}

The physics definition of a graph makes it easier to recast TAG in terms of the graph grammars in \citet{marcolli2015graph}. While it was possible to describe the production rules of TAG in terms of the combinatorial definition of graphs, the fact that tree-adjoining must be ultimately  viewed as an insertion-elimination operation introduces an unwelcome degree of context-dependence to the formulation. For every label at which adjunction may occur, there needs to be a distinct production rule for every possible degree and every possible combination of labels that the neighbours of this node may have, in order to account for all possible insertions. When using the physics definition, the production rules may be defined in terms of the corollas; in this case, we still need a distinct production rule for each possible valence but the labels of the neighbouring vertices no longer need to be considered.\footnote{For an example of this, see \S 3 of \citet{marcolli2015graph}.}

\subsection{Linguistic implications}\label{lingimp}

A striking benefit of this model is that null-adjoining constraints are \textit{automatically} granted based on whether or not an edge is allowed to be split into two half edges. If it is restricted from being two half-edges, then adjoining is  not possible. If it is required to split, then adjoining is required. This leads into another immediate linguistic payoff: feature-TAG.

In feature-TAG \cite{vijay1988feature}, the argument is made that tree adjoining is required when there is a featural mismatch in an elementary tree at a given node, e.g. it has [+WH] from above but [-WH] from below. Then in order to resolve this, the insertion of an auxiliary tree which has [+WH] in the upper node and [-WH] in the lower node would fix this, because the two [+WH] would match as well as the two [-WH], resolving the fact that originally the elementary tree had [+WH] and [-WH] in the same node. 


\begin{example}\label{ftag}
    To form the question ``What do you think Elizabeth lost?", we can use the following single elementary tree $T$ (left) and auxiliary tree $S$ (right):  

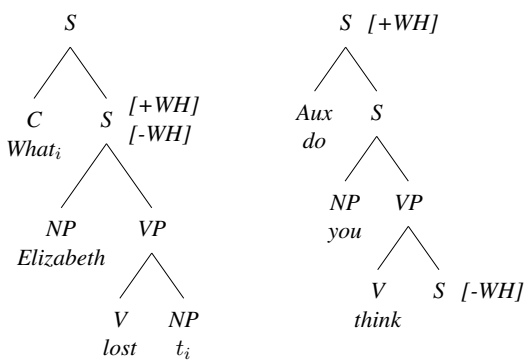
\[
\scalebox{.8}{\begin{forest}{nice empty nodes}
        [S [C \\What$_i$][S,label={[align=left]right:{[+WH]}\\{[-WH]}} [NP \\Elizabeth][VP [V\\lost][NP\\$t_i$]]]]
    \end{forest}  }
\hspace{1cm}
\scalebox{.8}{\begin{forest}{nice empty nodes}
        [S,label={[align=left]right:{[+WH]}} [Aux \\do][S [NP \\you][VP [V\\think][S,label={[align=left]right:{[-WH]}}]]]]
    \end{forest}  }
\]


In the elementary tree, there is a featural mismatch in the lower S node, because above this S node ``What" dictates it is [+WH], but below this S node there is no such lexical time/syntactic feature, meaning the lower feature is [-WH]. This featural mismatch prompts the insertion of $S$ into $T$, yielding the following well-formed tree with no featural mismatches:

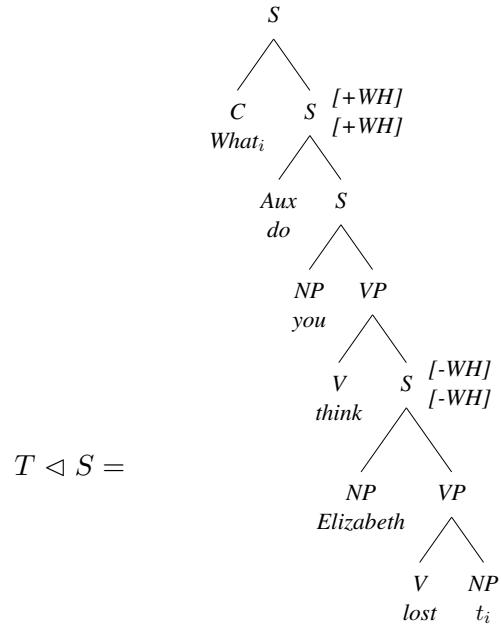
\begin{center}
    
$T \ins S=$\hspace{1cm}\raisebox{6cm}{\scalebox{.8}{\begin{forest}{nice empty nodes}
        [S [C \\What$_i$][S,label={[align=left]right:{[+WH]}\\{[+WH]}} [Aux \\do][S [NP \\you][VP [V\\think][S,label={[align=left]right:{[-WH]}\\{[-WH]}} [NP \\Elizabeth][VP [V\\lost][NP\\$t_i$]]]]]]]
    \end{forest}  }}
    \end{center}

\end{example}

Implementation of feature TAG follows easily from the half-edge model of TAG.\footnote{Note that this implementation is based upon labeling half-edges---this is an inherent component of the physics definition, so no additional functionality is being added to the system.} We can think about the half edges as carrying the feature labels, so an original $\alpha$ node's parent edge is split in half exactly when the upper half edge has a [+F] and the lower half edge is labeled by [-F] (or vice-versa). Then the auxiliary tree's upper half edge coming out of the root must carry a [+F] and the lower half edge, representing a leaf location, must carry a [-F]. After adjoining the two sets of two half edges will each have featural match. Below, we implement example (\ref{ftag}) with the physics-based TAG graphs.

\begin{example}
   The trees $T$ and $S$ are, respectively,
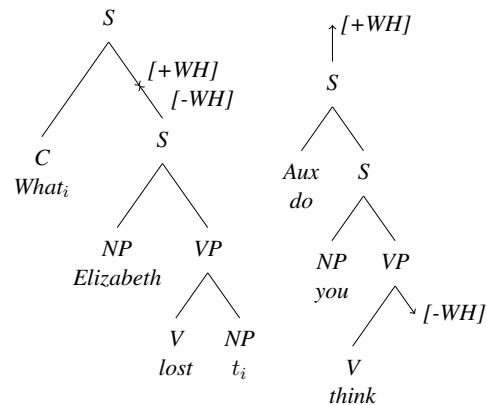
\begin{center}
\scalebox{.8}{\begin{forest}{nice empty nodes}
        [S [C \\What$_i$][,label={[align=left]right:{[+WH]}\\\quad{[-WH]}},l*= .5, edge = {->}[,no edge] [S,l*= .5, edge = {<-} [NP \\Elizabeth][VP [V\\lost][NP\\$t_i$]]]]]
    \end{forest}    }
\scalebox{.8}{\begin{forest}{nice empty nodes}
        [,label={[align=left]right:{[+WH]}},l*= .5[S, edge = {<-}  [Aux \\do][S [NP \\you][VP [V\\think][,l*= .5,edge = {->},label={[align=left]right:{[-WH]}}]]]]]
    \end{forest}    }
\end{center}
The insertion of $S$ into $T$ follows from the feature mismatch of the two half-edges---and this is resolved with the insertion:
 \begin{center}
$T \ins S=$\hspace{1cm}\raisebox{4cm}{\scalebox{.8}{\begin{forest}{nice empty nodes}
        [S [C \\What$_i$][,label={[align=left]right:{[+WH]}\\ \quad{[+WH]}},l*= .5, edge = {->}[,no edge][S, edge = {<-}, l*=.5 [Aux \\do][S [NP \\you][VP [V\\think][,label={[align=left]right:{[-WH]}\\ \quad{[-WH]}},l*= .5, edge = {->} [,no edge][S,l*= .5, edge = {<-} [NP \\Elizabeth][VP [V\\lost][NP\\$t_i$]]]]]]]]]
    \end{forest}  }}
    \end{center}
\end{example}

\subsection{Mathematical implications}

The Lie-algebraic structure provides many further avenues of exploration. The fact that we have managed to define graded and connected Lie algebra structures using TAG is convenient for future investigation of the universal enveloping algebra, since the universal enveloping algebra is graded, connected and furthermore commutative as a Hopf algebra. A graded connected commutative Hopf algebra is known to be free as an algebra (in other words, it is isomorphic to a polynomial algebra), thanks to a theorem of Hopf-Leray, which means that it can easily be implemented in computational algebra programs. If the Hopf algebra arises from a pre-Lie structure, then the coproduct can be defined in terms of the pre-Lie operator, which in our case has an intuitive combinatorial description in terms of trees. Indeed, any monomial over tree generators can be interpreted as a forest, so the Hopf algebra has a basis indexed by forests. All of these properties facilitate explicit calculations in the Hopf algebra.

As mentioned in the introduction, TAG insertion via nodes is only one type of graphical insertion. The other insertion type involves inserting into existing edges via splitting an edge into two with a node in the middle, and then connecting that node to the tree being inserted via another edge, as depicted in the following example.

\begin{example}
    Let $S,T$ be as previously, i.e.
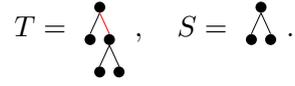
\[ T = \tinytree{[[][,edge = red[][]]]}, \quad  S = \tinytree{[[][]]}. \]
 Inserting $S$ into  $T$ at the red edge yields
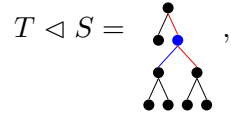
\[ T \ins S = \tinytree{[[][,edge = red, fill = blue [, edge = blue[][]][,edge = red[][]]]]},\]

where the single red edge has been split into two edges via the creation of the blue node and a new (blue) edge has been created connecting the root of $S$ to the new node.

\end{example}

This creation of two new edges allows the insertion operation to preserve the binary nature of the trees. It is worth recasting TAG insertion as an edge-insertion operation (e.g. above the labeled vertex, as the half-edges constructed in the physics formulation were) and  exploring the algebraic differences and implications of these two different types of TAG insertion. 

Another interesting perspective is due to a universal property of connected commutative Hopf algebras, which states that any such Hopf algebra is characterized by a unique nontrivial $1$-cocycle. For the Connes-Kreimer Hopf algebra of rooted nonplanar trees, which arises from the free pre-Lie algebra on one generator, this cocycle acts on forests by grafting the component trees to a common root. When the forest consists of exactly two trees, this provides a description of the fundamental Merge operation in \citet{marcolli2025mathematical}.

It would be interesting to examine the corresponding cocycles for the Hopf algebras obtained from the TAG Lie algebras. We expect them to be different from the grafting procedure since the pre-Lie operation based on TAG is not free.

\section{Conclusion}
In this paper, we formalized tree-adjoining grammars mathematically as Lie algebras where the pre-Lie operation was defined via the adjoining operation of TAG. In order to work with TAGs as mathematical objects, we demonstrated that representing the trees as graphs is only feasible in the Lie-algebra setting when we use the physics definition of graphs, allowing us to express single edges as two half-edges. This mathematical motivation also had striking linguistic payoffs, allowing various components of TAGs usually posited as additional constraints to be inherent via the nature of the graphs.

\section*{Acknowledgements}

We thank Robert Frank for his useful comments and suggestions. The third author is supported by NSF grant DMS-2104330.

\bibliography{custom}

\appendix

\section{Appendix: Proofs of Theorems}
\printProofs

\end{document}